\documentclass{article}

\usepackage{microtype}
\usepackage{graphicx}
\usepackage{subcaption}
\usepackage{booktabs} 

\usepackage{hyperref}

\usepackage[preprint]{icml2026}

\usepackage{amsmath}
\usepackage{amssymb}
\usepackage{mathtools}
\usepackage{amsthm}
\usepackage{tikz}
\usepackage{enumitem}

\usepackage{listings}
\usepackage[table]{xcolor}
\usepackage{colortbl}

\usepackage{graphicx}
\usepackage{booktabs}
\usepackage{tikz}
\usepackage[dvipsnames]{xcolor}
\usepackage{arydshln}

\definecolor{Red}{rgb}{0.768, 0.054, 0.054}
\definecolor{Green}{rgb}{0,0.4,0.7}
\hypersetup{
    colorlinks=true,
    citecolor=teal,
    linkcolor=Red,
    urlcolor=Green,
}

\usepackage[capitalize,noabbrev]{cleveref}

\theoremstyle{plain}

\theoremstyle{definition}

\theoremstyle{remark}

\usepackage[textsize=tiny]{todonotes}

\makeatletter
\renewcommand{\printAffiliationsAndNotice}[1]{%
  \global\icml@noticeprintedtrue%
  {\let\thefootnote\relax\footnotetext{$\dagger$ Equal advising. Correspondence to: kksan07@kaist.ac.kr}}%
}
\makeatother

\icmltitlerunning{Memory Transfer Learning: How Memories are Transferred Across Domains in Coding Agents}

\begin{document}

\twocolumn[
  \icmltitle{Memory Transfer Learning: \\ How Memories are Transferred Across Domains in  Coding Agents}



  \icmlsetsymbol{equal}{*}
  \icmlsetsymbol{co-advising}{$\dagger$}

  \begin{icmlauthorlist}
    \icmlauthor{Kangsan Kim}{KAIST}
    \icmlauthor{Minki Kang}{KAIST}
    \icmlauthor{Taeil Kim}{KAIST}
    \icmlauthor{Yanlai Yang}{NYU}
    \icmlauthor{Mengye Ren}{NYU,co-advising}
    \icmlauthor{Sung Ju Hwang}{KAIST,DeepAuto.ai,co-advising}
  \end{icmlauthorlist}
\begin{center}
    \rm\textsuperscript{1}KAIST
    \quad\textsuperscript{2}New York University
    \quad\textsuperscript{3}DeepAuto.ai
\end{center}
\begin{center}
\url{https://memorytransfer.github.io/}
\end{center}


  \icmlcorrespondingauthor{Kangsan Kim}{kangsan.kim@kaist.ac.kr}

  \icmlkeywords{Machine Learning, ICML}

  \vskip 0.3in
]



\printAffiliationsAndNotice{}  

\begin{abstract}
Memory-based self-evolution has emerged as a promising paradigm for coding agents. However, existing approaches typically restrict memory utilization to homogeneous task domains, failing to leverage the shared infrastructural foundations, such as runtime environments and programming languages, that exist across diverse real-world coding problems.
To address this limitation, we investigate \textbf{Memory Transfer Learning} (MTL) by harnessing a unified memory pool from heterogeneous domains. We evaluate performance across 6 coding benchmarks using four memory representations, ranging from concrete traces to abstract insights.
Our experiments demonstrate that cross-domain memory improves average performance by 3.7\%, primarily by transferring meta-knowledge, such as validation routines, rather than task-specific code. 
Importantly, we find that abstraction dictates transferability; high-level insights generalize well, whereas low-level traces often induce negative transfer due to excessive specificity.
Furthermore, we show that transfer effectiveness scales with the size of the memory pool, and memory can be transferred even between different models. 
Our work establishes empirical design principles for expanding memory utilization beyond single-domain silos.
\end{abstract}

\begin{figure}[t!]
    \centering
    \includegraphics[width=\linewidth]{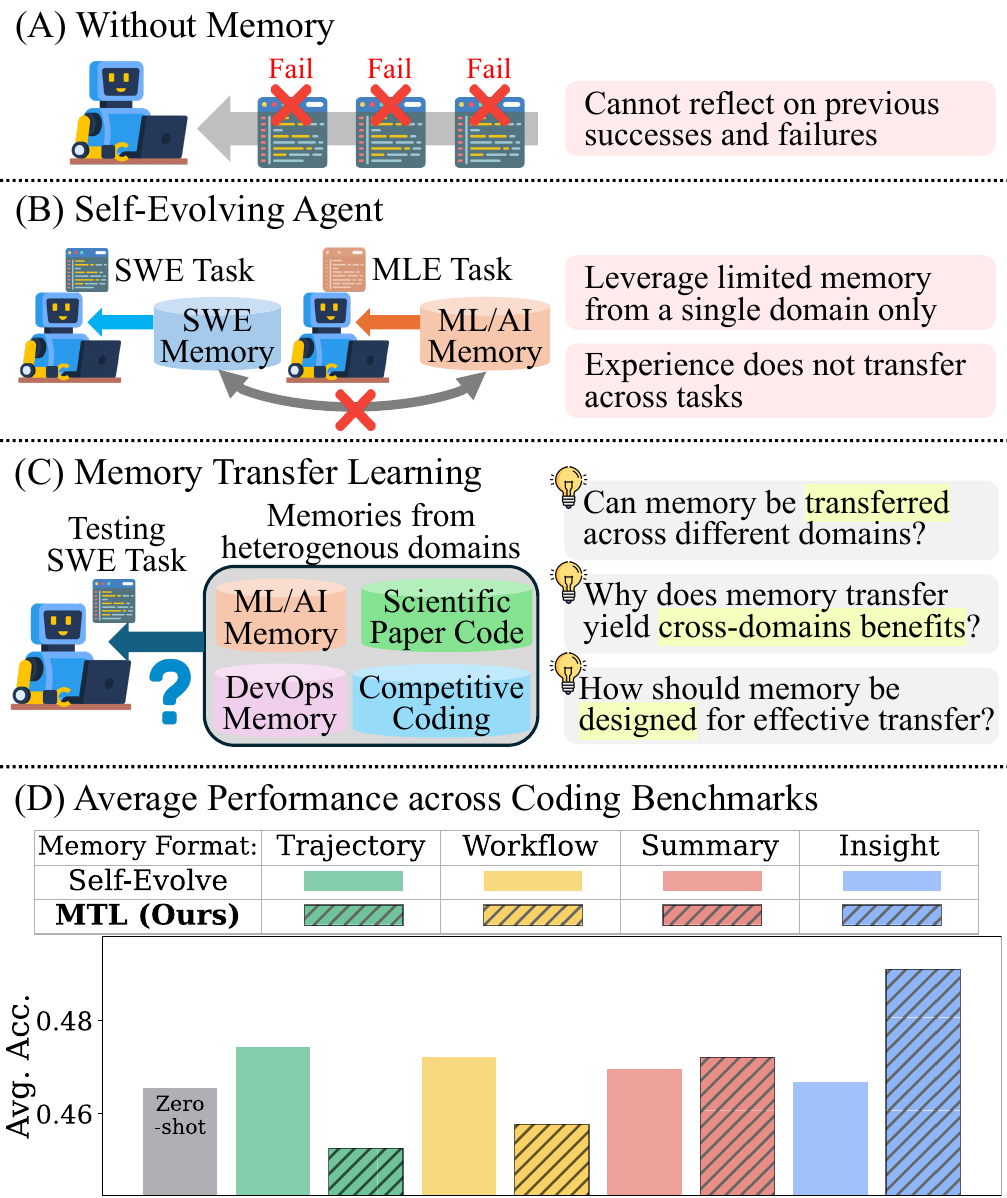}
    \caption{\textbf{Conceptual overview of Memory Transfer Learning.} Unlike (A) memory-less agents or (B) single-domain self-evolving agents, (C) our approach utilizes a shared memory pool from heterogeneous coding tasks. 
    (D) In the evaluation on diverse benchmarks, MTL outperforms a self-evolving approach. 
    }
    \label{fig:concept_figure}
    \vspace{-0.1in}
\end{figure}

\section{Introduction}
As performance gains from scaling training data in language models begin to plateau, self-evolution, which leverages prior inference outcomes to enhance future performance without additional supervision, has emerged as a promising paradigm for advancing model capabilities in agents~\cite{survey-selfevolve, survey2-selfevolve}. 
Memory plays a central role in self-evolving agents by enabling the extraction of reusable workflows and transferable insights from past inferences and their application to subsequent tasks~\citep{synapse, awm, reasoningbank}.
In coding agents, the memory is instantiated as code snippets, experiential knowledge including planning and debugging traces, or general programming principles~\citep{sweagent}.
Leveraging this knowledge allows agents to reference successful solution patterns in similar tasks, thereby reducing reasoning overhead, while also avoiding unnecessary failure actions in long-horizon code editing through adherence to accumulated procedural and strategic guidance, such as small-step modification heuristics and verification routines.

While memory-augmented coding agents have shown promise~\citep{reasoningbank}, existing approaches mostly restrict memory generation and retrieval to the same domain, typically within the same benchmark as illustrated in~\autoref{fig:concept_figure} (B).
However, in real-world scenarios, coding agents must handle a wide spectrum of programming problems, ranging from repository-level software engineering tasks~\citep{swebench} and machine learning model development~\citep{mlgym-bench, paper2code} to function-level competitive coding~\citep{livecodebench}.
Despite this diversity, these tasks share a common underlying infrastructure, including runtime environments (e.g., Linux shells), programming languages, and cross-file dependency stacks.
Current approaches that restrict memory utilization to a single domain fail to leverage this shared foundation, thereby preventing agents from exploiting a substantially richer memory pool derived from heterogeneous domains.
We posit that such cross-domain memories can provide valuable guidance, often more effective than those extracted solely from the same domain, by offering transferable knowledge applicable to new problems (\autoref{fig:concept_figure} (C)).

Some prior works have explored the construction of a large unified memory pool spanning multiple task types, providing initial evidence that general reasoning experiences can support software engineering tasks~\citep{agentkb}.
However, this line of works leaves several key research questions for a practical deployment unresolved, as in~\autoref{fig:concept_figure}:
 \vspace{-0.05in}
 \begin{list}{}{\setlength{\leftmargin}{1em} \setlength{\rightmargin}{1em}} 
\item \textit{\textbf{RQ1.} Does memory from heterogeneous domains improve the performance of coding agents?}
\item \textit{\textbf{RQ2.} Why do transferred memories yield benefits across different domains?}
\item \textit{\textbf{RQ3.} Which factors in memory transfer learning most influence transfer effectiveness?} \end{list}
\vspace{-0.05in}

\begin{figure*}[ht]
    \centering
    \includegraphics[width=\textwidth]{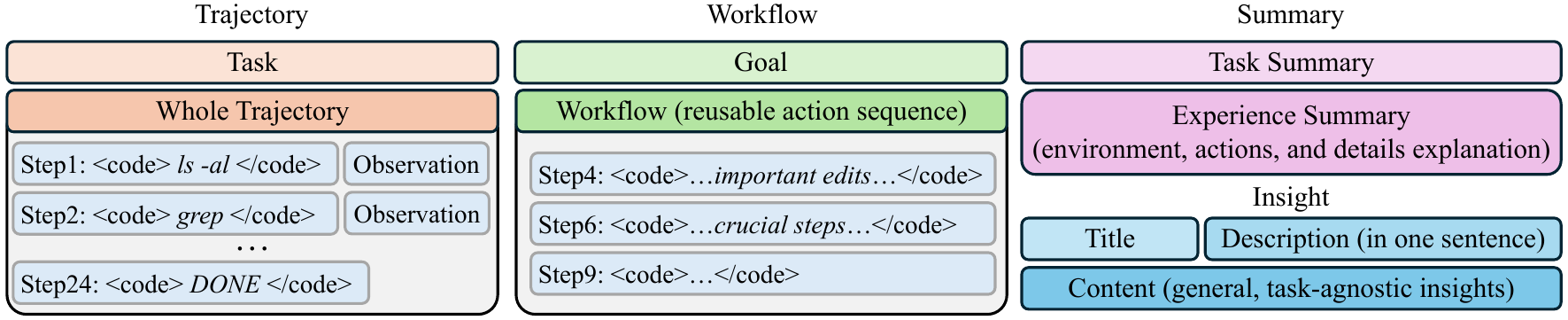}
    \vspace{-0.15in}
    \caption{\textbf{Illustrative examples of four memory formats.} We utilize Trajectory, Workflow, Summary, and Insight formats to analyze how different levels of information abstraction affect cross-task transferability.}
    \label{fig:memory_example}
    \vspace{-0.1in}
\end{figure*}

To address these open questions, we conduct a systematic investigation of \textbf{Memory Transfer Learning} across heterogeneous domains in coding agents and derive several core findings on its mechanisms and effects.
We first generate memories for each task using four different formats commonly adopted in prior works: Trajectory~\citep{synapse}, Workflow~\citep{awm}, Summary~\cite{reflexion}, and Insight~\citep{reasoningbank}, as illustrated in \Cref{fig:memory_example}.
We then evaluate coding agent performance in zero-shot setting and under Memory Transfer Learning.
The results demonstrate that Memory Transfer Learning can provide effective and transferable knowledge, \textbf{improving 3.7\% of average scores of 6 coding benchmarks.}

Our analysis yields three core findings into the mechanism of memory transfer.
\textbf{First, cross-domain significantly improve the performance of coding agents.} Although existing self-evolving methods often overlook out-of-domain memories, our results suggest that effective memory utilization should incorporate all past experiences, including those from different domains, to enhance agent performance, as shown in \autoref{fig:concept_figure} (D).
\textbf{Second, the primary transferable value lies in meta-knowledge.}
Through qualitative analysis, we find that cross-task benefits stem not from task-specific code content but from operational know-how, such as preventing execution failures under environment constraints and task-solving routines that prioritize structural and interface inspection followed by strict validation procedures.
\textbf{Third, abstraction dictates transferability.}
By quantifying the abstraction level of each memory format, we discover a positive correlation between high-level abstraction and transfer effectiveness.
Highly abstract memories, such as Insights, become task-agnostic and generalizable.
In contrast, low-abstraction memories like Trajectories retain excessive task-specific details that can distract the agent, confirming that raw execution traces are less suitable for cross-task transfer.
Furthermore, we provide additional insights into Memory Transfer Learning, including why negative transfer occurs, how the performance gain of MTL scales with a larger memory pool and more domains, and the potential for transferring memories across different models.


In conclusion, this work presents a first holistic investigation of Memory Transfer Learning.
Importantly, through extensive evaluation on six coding benchmarks, we show that existing agents' memory usage methods, which focus on a homogeneous domain, are limited, and that there is significant room for improvement by leveraging memories from heterogeneous domains.
We hope this study expands the scope of memory utilization beyond single-domain settings and stimulates further research on how to effectively leverage memory in self-evolving agents, ultimately leading to more capable coding agents.

\section{Related Work}
\subsection{Coding Agents}
As LLM have demonstrated strong capabilities in code generation~\cite{codellama, qwen2.5coder, deepseekcoder2}, researchers have developed LLM-based coding agents that interact with programming environments such as bash shells~\cite{harbor, openhands} through diverse systematic designs, and have evaluated them across a wide range of coding tasks. At the early stage of coding agents, they target function-level code generation tasks~\cite{autocodebench, livecodebench, evoeval} in a single file. AlphaCodium~\cite{alphacodium} proposed a flow engineering in code generation which iteratively run reasoning, generation, ranking, and debugging. LDB~\cite{ldb} introduced a novel debugging framework with language models that leverage runtime execution information for function-level code generation. Beyond a single file level editing, CodeAgent~\cite{codeagent}, RepoAgent~\cite{repoagent}, RLCoder~\cite{rlcoder} address repository-level code modification tasks~\cite{swebench, terminalbench}. Furthermore, code agents targeting domain-specific tasks, such as Paper2Code~\cite{paper2code} for code generation for ML paper replication tasks and BixbBench~\cite{bixbench} for computational biology related tasks.

\subsection{Memory-based Self-Evolving Agents}
Self-evolving agents~\cite{flex} leverage past experiences by reusing successful solution patterns in similar tasks and avoiding previously encountered erroneous actions. To manage these experiences effectively, existing memory-based self-evolving agents~\cite{worldmm, ma-egoqa} primarily focus on mechanisms for memory generation and retrieval during interactions with environments.~\cite{memp, swe-exp} AWM~\cite{awm} proposed memory utilization through the collection of common workflows in web agents, while ReasoningBank~\cite{reasoningbank} extracts helpful insights from trajectories via test-time scaling. Dynamic Cheatsheet~\cite{dynamic_cheatsheet} constructs evolving memories that encode reusable strategies and insights, and ReMe~\cite{reme} presents a holistic framework from memory generation to retrieval and memory refinement. MemEvolve~\cite{memevolve} further introduces system-level evolution through meta-evolution in memory agents. However, existing memory-based self-evolving agents are primarily evaluated within the same benchmark or task domain, overlooking the potential value of memories generated from other task domains that may be highly beneficial to agent performance.

\subsection{Transfer Learning}
Transfer Learning~\cite{survey_tl} has been extensively studied as the reuse of knowledge acquired in a source domain to improve performance in a target domain. Traditional approaches mainly rely on parametric adaptation through model updates~\cite{ulmfit, adapters}. With the emergence of LLMs demonstrating strong generalization capabilities, recent work has increasingly explored non-parametric knowledge transfer mechanisms. In-context learning~\cite{icl_survey, rethinking_icl, videoicl}, as a representative paradigm, shows that LLMs can reuse knowledge provided in the context at inference time. In the agent setting, knowledge is instead generated by the model itself in the form of memory and transferred across tasks. AgentKB~\cite{agentkb} introduces a framework for managing and leveraging a unified memory pool across multiple task domains. However, it does not provide a deeper analysis of the underlying mechanisms of memory transfer, including which forms of knowledge are transferable and how transfer-oriented memories should be generated in contrast to in-domain knowledge. Moreover, prior work typically constructs unified memory spaces across heterogeneous environments, such as general reasoning, web interaction, and coding, thereby missing the opportunity to exploit coding-specific shared principles that are unique to programming tasks.

\section{Memory Transfer Learning}
We introduce Memory Transfer Learning, which leverages memories generated from heterogeneous tasks with target tasks in coding environments. 
In the following sections, we first describe how we generate and retrieve memory, and which benchmarks we use to evaluate the performance. 

\subsection{Method}
To investigate the impact of memory on the agent, we design a simple memory-based coding agent with a two-stage memory utilization process: memory generation and memory retrieval. Memory generation is performed offline with results saved prior to memory transfer learning, while memory retrieval is executed for each query during the inference.

\subsubsection{Memory Generation}
Before memory generation, we first run inference the agent across all benchmarks and gather the resulting trajectories as sources for memory construction. Inference results consist of the given task $t$ and multiple steps of reasoning $r$, action $a$, observation $o$, thus the full inference history $H$ is denoted as $H = (t, [(r_1, a_1, o_1), \ldots, (r_n, a_n, o_n)])$ with task $t$. Based on these results, we construct four types of memory representations, defined by categorizing memory schemes from existing self-evolving agents into representative formats. We employ LLM-based judge to assess whether each inference attempt is successful or failed, and use different memory generation prompts for each case, following previous work~\cite{reasoningbank, reme}. Detailed descriptions for each memory format is as follows. The structure illustration for each format is shown in~\autoref{fig:memory_example}, and prompts used in memory generation are in \autoref{appendix:prompts}.

\textbf{Trajectory} In this memory representation, we concatenate all commands and codes called by the agent $a_i$ and their execution results $o_i$ from $H$ without reasoning sentences $r_i$, and save it with the source task $t$. Trajectory memory $M_T$ can defined as $M_T = (t, [(a_1, o_1), \ldots, (a_n, o_n)])$. This contains detailed information of task solving experience even with failed steps. Also the agent can implicitly estimate the expected execution results of certain actions by referring observations of similar commands in this memory.   

\textbf{Workflow} In order to focus only on meaningful code snippets in the entire trajectory, this memory representation is generated by extracting reusable workflow from the trajectory.~\cite{awm} Specifically, we provide $H$ to LLM and ask to generate a goal of workflow $g$ and extract meaningful actions $a$ to achieve the goal. Therefore, workflow memory $M_W$ denotes as $M_W = (g, [a_i, a_j, \ldots, a_k])$. By restoring a subset of the action and observation history, Workflow is much shorter than Trajectory which leads to less danger of distractions from unrelated information. 

\textbf{Summary} One key principle in leveraging memory is to follow the successful actions and reflect failures from previous inference, however, raw code commands and observations do not provide explicit information about analysis why the agent succeeds or fails and the findings from the history. Thus, for Summary memory, we prompt LLM to summarize the task, environment, actions, results, and analysis on why this inference succeeds or fails from the given trajectory. In detail, LLM generates a summary of task $s_t$ and one paragraph of experience summary $s_e$ from the trajectory, which is represented as $M_S = (s_t, s_e)$ for Summary memory $M_S$. 

\textbf{Insight} We can reasonably expect that memory should be generalized to be easily adapted to different tasks, and as the most general memory representation, we employ the Insight memory format. Following the memory design ReasoningBank~\cite{reasoningbank}, Insight $M_I$ consist of three parts: title $i_t$, description $i_d$, content $i_c$, represented as $M_I = (i_t, i_d, i_c)$. In the content of this memory item, we prompt LLM to write insights on why this task is successfully accomplished without mentioning specific files or details. Additionally, we explicitly instruct LLM to generate generalizable insights for future similar tasks. 

\begin{table*}[t]
    \centering
    \caption{\textbf{Evaluation results of Memory Transfer Learning.} We report Pass@3 scores across multiple benchmarks. MTL consistently improves performance over the zero-shot baseline across models. Among memory types, Insight achieves the highest average performance.}
    \label{tab:full_results}
    \resizebox{\linewidth}{!}{
    \begin{tabular}{l cc cc cc c}
    \toprule
    & \textbf{LiveCodeBenchv6} & \textbf{Aider-Polyglot} & \textbf{SWEBench-Verified} & \textbf{TerminalBench2} & \textbf{ReplicationBench} & \textbf{MLGym-Bench} & \textbf{Avg.} \\
    \midrule
    \midrule
    \rowcolor{gray!10}
    \textbf{GPT-5-mini} & & & & & & & \\
    Zero-shot & 0.910 & 0.470 & 0.730 & 0.315 & 0.111 & 0.667 & 0.523 \\
    MTL (T) & 0.940 & 0.490 & 0.770 & 0.270 & 0.122 & 0.583 & 0.534 \\
    MTL (W) & 0.920 & 0.470 & 0.770 & 0.348 & 0.111 & 0.583 & 0.538 \\
    MTL (S) & 0.930 & 0.460 & 0.760 & 0.371 & 0.133 & 0.667 & \underline{0.546} \\
    MTL (I) & 0.930 & 0.470 & 0.770 & 0.360 & 0.189 & 0.750 & \textbf{0.560} \\
    \rowcolor{blue!10}
    $\Delta$ & +2.0\% & 0.0\% & +4.0\% & +4.5\% & +7.8\% & +8.3\% & +3.7\% \\
    \noalign{\vskip 0.25ex}\cdashline{1-8}\noalign{\vskip 0.75ex}
    \rowcolor{gray!10}
    \textbf{DeepSeek V3.2} & & & & & & & \\
    Zero-shot & 0.930 & 0.590 & 0.530 & 0.337 & 0.267 & 0.583 & 0.542 \\
    MTL (I) & 0.940 & 0.580 & 0.590 & 0.393 & 0.278 & 0.667 & \textbf{0.568} \\
    \rowcolor{blue!10}
    $\Delta$ & +1.0\% & -1.0\% & +6.0\% & +5.6\% & +1.1\% & +8.3\% & +2.6\% \\
    \noalign{\vskip 0.25ex}\cdashline{1-8}\noalign{\vskip 0.75ex}
    \rowcolor{gray!10}
    \multicolumn{3}{l}{\textbf{Qwen3-Coder-480B-A35B-Instruct}} & & & & & \\
    Zero-shot & 0.800 & 0.460 & 0.590 & 0.292 & 0.211 & 0.583 & 0.483 \\
    MTL (I) & 0.810 & 0.480 & 0.620 & 0.326 & 0.211 & 0.583 & \textbf{0.501} \\
    \rowcolor{blue!10}
    $\Delta$ & +1.0\% & +2.0\% & +3.0\% & +3.4\% & 0.0\% & 0.0\% & +1.8\% \\
    \bottomrule
    \end{tabular}
    }
\vspace{-0.1in}
\end{table*}

\subsubsection{Memory Retrieval}
\textbf{Memory Pool Construction} After finishing memory generation for all benchmarks, we construct the heterogeneous-domain memory pool to experiment memory transfer learning. 
We gather memories from all benchmarks except the testing benchmark for each memory format. In formal notation, the memory pool $\mathcal{P}$ used for memory transfer learning in evaluating benchmark $B_i$ with memory type $\tau$ is $\mathcal{P}_{\tau}(B_i) = \{ M_\tau^{(k)} \mid t^{(k)} \not\in B_i \}_{k=1}^{N_i}$. When constructing the memory pools, we index each memory by extracting embedding features using a textual embedding model and store the features with the memories. 

\textbf{Memory Retrieval} In the inference stage, we retrieve $N$ relevant memories for each task from the memory pool correspond to the testing model, benchmark, and memory type, and provide retrieved memories into the system prompt of the coding agent at the beginning of the inference. In detail, we generate the embedding feature of current task and measure the cosine similarity between task embedding feature and memory embedding features. Finally, we select the final retrieved memories by top-$N$ sampling with the highest similarity scores.  

\subsection{Experimental Details}
\subsubsection{Datasets}
We evaluate the coding agents with different memory utilization methods with 6 different coding benchmarks. For competitive and function-level programming tasks, we use Aider Polyglot~\cite{aider-polyglot} and LiveCodeBenchv6~\cite{livecodebench}. For repository-level coding tasks, we employ SWE-Bench Verified~\cite{swebench} and Terminal Bench2~\cite{terminalbench}. We also evaluate the performance on domain-specific code generation benchmarks, such as ReplicationBench~\cite{replicationbench} for scientific knowledge grounding code generation and MLGym-Bench~\cite{mlgym-bench} for machine learning research tasks. For all benchmarks, we randomly sample 100 tasks if the total number of sample is over 100, and we evaluate task success using evaluation protocol of each benchmark and report performance in terms of Pass@3.

\subsubsection{Additional Details}
We adopt gpt-5-mini model for every LLM usage from generating memories, base model for coding agent, to a LLM judge. We also exploit mini-swe-agent~\cite{sweagent} for coding agent, harbor~\cite{harbor} for evaluation platform, and text-embedding-3-small model of OpenAI for text embedding extraction. In the memory retrieval stage, we select three memories for each query ($N = 3$). In querying Trajectory memory, we use embedding similarities between target task and tasks in the memories, since both query and memories have the \textit{Task} information. In querying other memories (Workflow, Summary, and Insight), which do not have \textit{Task} information, we ask the model to write 4-5 sentences of coding plan to solve the given task and use the plan as the query.  

\section{Experimental Results and Analysis}
We now present and analyze the results of our experiments, and introduce academic findings on characteristics of memory transfer learning based on analysis in diverse aspects. 

\begin{table}[t!]
    \renewcommand{\arraystretch}{1.1}
    \centering
    \caption{\textbf{Pass@3 comparison with self-evolving methods.} LCB, SWEB, and RepliB denote LiveCodeBenchv6, SWEBench-Verified, and ReplicationBench, respectively.}
    \label{tab:self_evolve}
    \vspace{-0.05in}
    \resizebox{0.99\linewidth}{!}{
    \begin{tabular}{l c ccc c}
    \toprule
    \textbf{Method} & \textbf{\#Memories} & \textbf{LCB} & \textbf{SWEB} & \textbf{RepliB} & \textbf{Avg.} \\
    \midrule
    \midrule
    Zeroshot & - & 0.910 & 0.730 & 0.111 & 0.584 \\
    ReasoningBank & 97 & 0.920 & 0.750 & 0.133 & 0.601 \\
    AgentKB & 5,899 & 0.920 & 0.720 & 0.200 & \underline{0.613} \\
    MTL (Ours) & 431 & 0.930 & 0.770 & 0.189 & \textbf{0.630} \\
    \bottomrule
    \end{tabular}
    }
\vspace{-0.15in}
\end{table}

\subsection{Overall Performance of MTL}

\subsubsection{Main Results}
\label{sec:main_result}
The results of the coding agent with Memory Transfer Learning across six coding benchmarks are shown in \autoref{tab:full_results}. Performance of Memory Transfer Learning significantly improves the performance compared to zero-shot setting, in particular, when Insight memories are transferred, the agent achieves more than 4.0\% (up to 8.3\%) performance gains on four benchmarks. 
These results highlight that transferable knowledge exists across different domains, and leveraging such knowledge is crucial for improving the performance of coding agents.
Moreover, these results are further validated across different models. We evaluate the effectiveness of Memory Transfer Learning on the DeepSeek V3.2~\cite{deepseek-v3.2} and Qwen3-Coder-480B-A35B-Instruct~\cite{qwen3, qwen3-coder-next} models, achieving average performance improvements of 2.6\% and 1.8\%, respectively.
Notably, these results indicate that our method is also beneficial for open-sourced models, highlighting the broad applicability of cross-domain memory transfer.

\subsubsection{Comparison with Self-Evolving Approaches}
We further evaluate Memory Transfer Learning against two representative self-evolving methods, ReasoningBank~\cite{reasoningbank} and AgentKB~\cite{agentkb}, on three benchmarks. Each model is evaluated over three runs, and we report Pass@3 scores to ensure the robust comparison.
As presented in \autoref{tab:self_evolve}, Memory Transfer Learning outperforms both self-evolving methods by +2.9\% and +1.7\%, respectively, demonstrating that its performance gain are substantial even relative to strong self-evolving baselines. ReasoningBank achieves the lowest average gain, as it only leverages a small number of in-domain memories and does not utilize cross-domain knowledge. In contrast, AgentKB leverages a large number of out-of-domain memories (from general reasoning tasks), but still underperforms our method despite using around 5.8k memories. Notably, Memory Transfer Learning uses only 431 memories in the memory pool, yet achieves the highest average performance. This demonstrates both the effectiveness and efficiency of our approach compared to existing self-evolving methods. 

\begin{tikzpicture}
\node[draw, rounded corners=5pt, fill=green!10,
minimum width=\columnwidth, minimum height=1cm,
align=left, anchor=west, text width=\columnwidth-0.45cm,
inner xsep=5pt, inner ysep=6pt] at (0,0) 
{\textbf{Core Finding 1.} Memory Transfer Learning significantly improves coding agent performance and outperforms self-evolving methods in effectiveness and efficiency.};
\end{tikzpicture}

\begin{figure}[t!]
    \centering
    \includegraphics[width=\linewidth]{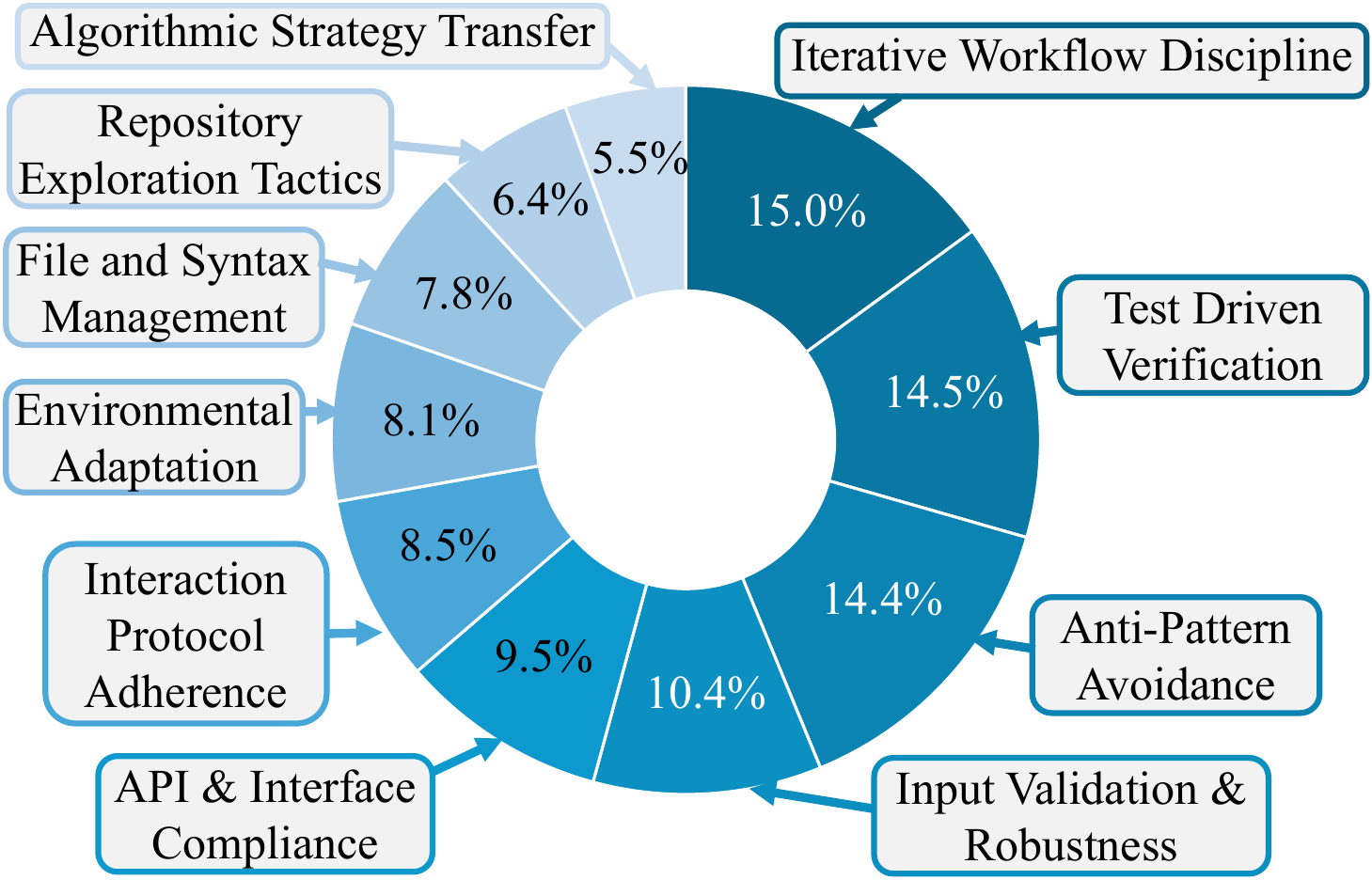}
    \caption{\textbf{Breakdown of Memory Transfer Contribution.} Transferred memory mainly contributes through meta-knowledge.}
    \label{fig:donut_figure}
    \vspace{-0.3in}
\end{figure}
\begin{table*}[t!]
    \centering
    \caption{\textbf{Case Study: Zero-shot vs. Memory Transfer Learning with Insight.} Transferred Insight from LiveCodeBench to SWE-Bench Verified provides meta-knowledge regarding strategic guidance (\textit{e.g.,} inline test validation), allowing the agent to succeed where the zero-shot baseline fails.}
    \label{tab:case_meta_memory}
    \vspace{-0.05in}
    \small
    \resizebox{\textwidth}{!}{
        \begin{tabular}{p{0.08\linewidth}|p{0.34\linewidth}|p{0.5\linewidth}}
        \toprule
        \rowcolor{blue!5}
         \textbf{Model} & \textbf{Zero-shot} & \textbf{Memory Transfer Learning with Insight} \\
        \midrule
         \textbf{Memory} & (No Memory) & Memory Item 2 (Generated from LiveCodeBench):
         \newline
         \textbf{Title}: Create quick self-contained tests using an inline Python here-doc to validate fixes 
         \newline
         \textbf{Description}: When making small code fixes, write a minimal tests...
         \newline
        \textbf{Content}: Set up a short battery of tests that cover ...
        \\
         \midrule
         \textbf{Reasoning} & I will modify django/db/models/aggregates.py to \textcolor{red}{raise a FieldError} when an Aggregate's source expression contains a window (over) clause. This prevents creating an aggregate over a window expression ...
        & I will use \textcolor{blue}{Memory Item 2 (use an inline Python here-doc for safe, atomic edits and quick verification)} to modify django/db/models/sql/query.py. The goal is to ensure that when resolving aggregates we detect if any referenced annotation either is a subquery or contains a window expression ... \\
        \midrule
        \textbf{Result} & Fail & Success \\
         \bottomrule
        \end{tabular}
    }
    \vspace{-0.1in}
\end{table*}

\subsection{Mechanism of Memory Transfer Learning}

\subsubsection{How Does Memory Transfer Learning Benefit the Agents?}
\label{sec:how_memory_benefit}
To investigate the operational mechanisms of memory transfer learning, we inspect the inference outcomes using LLM and manual case studies. First, we collect the trajectories of the instances in which the agent fails in the zero-shot setting but succeeds when Memory Transfer Learning with Insight memory is applied, and use GPT-5 to categorize how transferred memory contributes to successful task completion. As presented in \autoref{fig:donut_figure}, our analysis reveals that transferred memory primarily benefits agents by providing meta-knowledge rather than task-specific programming content. This meta-knowledge includes structured action workflow (e.g., inspect, edit, verify, submit), guardrails for compliance with external constraints (such as output formats, function signatures, and API contracts), and disciplined programming practices that discourage large one-shot refactors, blind overwrites, and brittle hardcoding. 

Transferred memory further promotes risk-controlled editing through minimal patch strategies, self-generated verification when official tests are unavailable, and safe interaction with execution environments by anticipating tool-chain and infrastructure failures. By supplying procedural guidance on how to act and how to safely interact with the execution and testing environment, Memory Transfer Learning enables agents to follow stable inference patterns and significantly reduces failure cases caused by infrastructure-level errors. 
On the other hand, \textit{Algorithmic Strategy Transfer} accounts for only 5.5\% of the total gains in \autoref{fig:donut_figure}, suggesting that the direct transfer of specific programming knowledge or algorithms is limited in our setting.

\subsubsection{Case Study: Zero-shot vs. MTL}
The effect of Memory Transfer Learning and transferred meta-memory are also shown in the case study. In \autoref{tab:case_meta_memory}, we compare the inference outcomes between zero-shot and memory transfer learning tested on one instance of SWEBench-Verified. In zero-shot setting, model naively solve the task by simply raising an error and eventually fail the test. However, with memory transfer learning, retrieved Insight memory generated from LiveCodeBench provides behavior knowledge about testing with an inline Python here-doc to validate fixes, and the agent follows the guideline of it and successfully completes the task. This case study highlights the practical impact of transferred meta-memory in enabling successful task completion. 

\begin{tikzpicture}
\node[draw, rounded corners=5pt, fill=green!10,
minimum width=\columnwidth, minimum height=1cm,
align=left, anchor=west, text width=\columnwidth-0.45cm,
inner xsep=5pt, inner ysep=6pt] at (0,0) 
{\textbf{Core Finding 2.} Transferable knowledge exists across distinct task types, and its primary form is meta-memory encoding procedural and behavioral guidance, not domain-specific knowledge.};
\end{tikzpicture}
\vspace{-0.2in}

\begin{figure*}[t!]
  \centering
  \begin{minipage}[t]{0.95\textwidth}
    \centering
    \includegraphics[width=\textwidth]{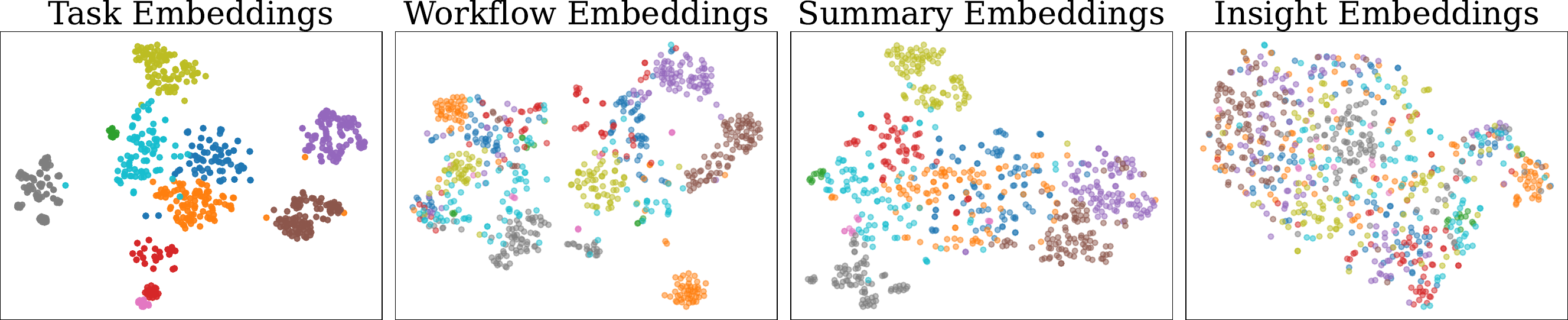}
  \end{minipage}
  \vspace{-0.05in}
  \caption{\textbf{t-SNE Visualization of Memory Formats.} The leftmost plot shows task embeddings, followed by three different memory types to the right. Each color represents a specific benchmark used in experiments. While task and workflow embeddings are clustered within each domain, the insight embeddings are sparse and intermingled, reflecting their task-agnostic nature.}
    \label{fig:t-sne}
  \vspace{-0.1in}
\end{figure*}

\subsection{Impact of Memory Abstraction}
\subsubsection{Abstraction Level of Four Memory Types}
\begin{figure}[t!]
    \centering
    \vspace{-0.1in}
    \includegraphics[width=\linewidth]{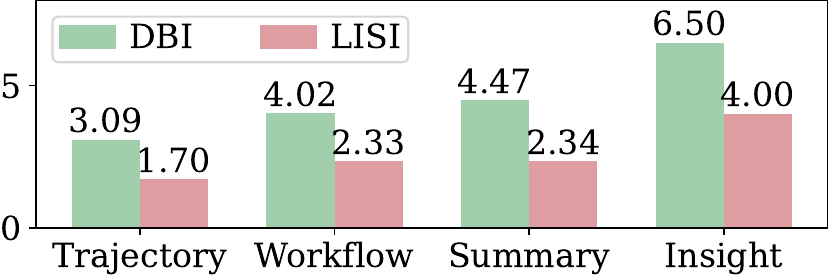}
    \vspace{-0.2in}
    \caption{\textbf{Embedding Distribution Analysis} DBI and LISI reveal weaker separation and stronger mixing with higher abstraction.}
    \label{fig:index_metric}
    \vspace{-0.2in}
\end{figure}

We adopt four memory representations in our experiments, each designed with a distinct level of abstraction. Trajectory and Workflow memories are less abstract and highly task-specific, containing raw command-level actions, whereas Summary and Insight memories are more abstract and generalized. These properties are clearly reflected in the embedding space visualizations shown in \autoref{fig:t-sne}. Task embeddings form benchmark-level clusters, while embeddings in the Insight space become increasingly sparse and intermingled across benchmarks, indicating that Trajectory and Workflow remain task-specific, whereas Summary and Insight exhibit greater generality. 
In addition, memory embedding distributions are quantitatively characterized using the Davies–Bouldin Index (DBI) and the Local Inverse Simpson’s Index (LISI), as shown in \autoref{fig:index_metric}. The increasing DBI values indicate progressively weaker benchmark-level cluster separation, while the increasing LISI values indicate stronger local benchmark mixing, quantitatively supporting the transition from task-specific to generalized memory.

\subsubsection{Correlation between Abstraction and Transfer Effectiveness}
In \autoref{tab:full_results}, under the memory transfer learning setting, the Insight format achieves the best performance, followed by Summary, Workflow, and Trajectory, while all MTL variants outperform the zero-shot baseline. As discussed in \autoref{sec:how_memory_benefit}, transferred memories primarily provide general meta-knowledge, whereas implementation-specific details from unrelated tasks may distract the agent. Consequently, more abstract and generalized memory representations tend to yield higher transfer effectiveness.

\subsubsection{Isolating the Effect of Memory Abstraction}
To isolate the effect of abstraction, we compare two groups of memories within the same representation (Insight): relatively task-specific and task-agnostic memories. 
Specifically, we prompt an LLM to infer the original task solely from each Insight memory and measure the similarity between the inferred task and ground-truth task. 
Higher similarity indicates that the memory retains more task-specific information, while lower similarity indicates that it is more abstract and task-agnostic.
Based on this measure, we partition the memories into the top 30\% (task-specific) and bottom 30\% (task-agnostic). This controlled setup allows us to evaluate the effect of abstraction while keeping the memory format fixed.
In \autoref{tab:causal_analysis}, we observe that even within the same memory format, task-agnostic memories consistently outperform task-specific ones. 
This provides evidence that abstraction, rather than format itself, is a key factor driving transfer performance.
Furthermore, we present the formal modeling on the correlation of memory abstraction and transfer effectiveness in \autoref{appendix:formal_modeling}.

\begin{table}[t!]
    \renewcommand{\arraystretch}{1.1}
    \centering
    \vspace{-0.1in}
    \caption{\textbf{Memory Abstraction Effect} Task-agnostic insights outperform task-specific ones, highlighting abstraction as a key driver.}
    \label{tab:causal_analysis}
    \renewcommand{\arraystretch}{1.1}
    \resizebox{0.99\linewidth}{!}{
    \begin{tabular}{l ccc c}
    \toprule
    \textbf{Method} & \textbf{LCB} & \textbf{SWEB} & \textbf{RepliB} & \textbf{Avg} \\
    \midrule
    \midrule
    Task-specific Insights & 0.887 & 0.617 & 0.067 & 0.523 \\
    Task-agnostic Insights & 0.893 & 0.627 & 0.082 & \textbf{0.534} \\
    \rowcolor{gray!20}
    $\Delta$ & +0.6\% & +1.0\% & +1.5\% & +1.1\% \\
    \bottomrule
    \end{tabular}
    }
\vspace{-0.1in}
\end{table}

\begin{table*}[t!]
    \centering
    \caption{\textbf{Case Study: Memory Transfer Learning with Trajectory vs. Insight.} Trajectory memory transferred from MLGym-Bench illustrates that blidly following low-abstraction memory leads to execution errors due to task-specific command. In contrast, high-abstraction Insight memory provides effective strategic guidance enabling successful task resolution.}
    \label{tab:case_abstraction_level}
    \vspace{-0.05in}
    \small
    \resizebox{0.97\textwidth}{!}{
        \begin{tabular}{p{0.08\linewidth}|p{0.45\linewidth}|p{0.47\linewidth}}
        \toprule
        \rowcolor{blue!5}
         \textbf{Model} & \textbf{Memory Transfer Learning with Trajectory} & \textbf{Memory Transfer Learning with Insight} \\
        \midrule
         \textbf{Memory} & Memory Item 1 (Generated from MLGym-Bench): \
         \newline
         \textbf{Task}: You are given the train, validation and test sets of Housing Price Prediction dataset and a baseline...
        \newline
        
        \textbf{Trajectory}:
        \newline
         \scriptsize{
         \textbf{Agent}: \texttt{ls -la data \&\& nl -ba evaluate.py | sed -n '1,240p'}
         \newline
         \textbf{System}: \texttt{<returncode>0</returncode><output>total 464 drwxr-xr-x 2 root root 4096 Dec 25 18:50 .drwxrwxrw- 1 agent agent 4096 Jan 14 06:02 ...}
         \newline
         \textbf{Agent}: \texttt{cat <<'PY' > train\_predict.py ... ('onehot', OneHotEncoder(..., sparse=False)) ... PY}
         \newline
         \textbf{System}: \texttt{TypeError: OneHotEncoder.\_\_init\_\_() got an unexpected keyword argument 'sparse'}} ...
         & Memory Item 1 (Generated from MLGym-Bench):
         \newline
         \textbf{Title}: Inspect eval requirements, combine train+val, use robust preprocessing, and adapt to library API changes 
         \newline
         \newline
         \textbf{Description}: Before improving a baseline model, read the repository/evaluation expectations, train on all available labeled data, apply robust preprocessing for numeric and categorical ...
         \newline
         \newline
        \textbf{Content}: First inspect the code that loads labels and evaluates predictions so you know the exact submission format and metrics to optimize. For the final model, combine training ...  \\
         \midrule
         \textbf{Reasoning} & I'll start by inspecting the repository root... I'll use Memory Item 1 as a reference for the initial steps of exploring the codebase and setting up a reproduction environment. \newline 
         \textcolor{red}{(Blindly following the trajectory resulted in an error.)}
        & I will modify gradient\_boosting.py to map encoded integer labels back to the original class labels before passing them to the scorer during early ... \textcolor{blue}{I'll use Memory Item 2 (inspect/evaluate and adapt code) as guidance to carefully inspect and modify the code.} \\
        \midrule
        \textbf{Result} & Fail & Success \\
         \bottomrule
        \end{tabular}
    }
    \vspace{-0.1in}
\end{table*}

\subsubsection{Case Study: Trajectory vs. Insight}
We further validate the relationship between memory abstraction level and transferability through qualitative case studies. In \autoref{tab:case_abstraction_level}, we compare representative examples of memory transfer using Trajectory and Insight format. In the Trajectory transfer example, the agent blindly follows the exact commands-level instructions in the memory. This behavior is inherently risky, since implementation details often differ across tasks and environments, including programming languages, file structures, and execution pipelines. As a result, the transferred Trajectory memory acts as a brittle anchor, leading the agent to execute incompatible commands and ultimately causing runtime errors.

In contrast, the transferred Insight memory provides high-level behavioral guidance, such as prioritizing inspection of evaluation criteria and improving data utilization by merging training and validation sets. Rather than imposing concrete implementation details that may conflict with the new task, it supplies abstract procedural principles that guide the agent’s reasoning without constraining its adaptation process. As reflected in the reasoning trace, the agent internalizes these general coding practices while deriving task-specific implementation details, leading to successful task completion.

\begin{tikzpicture}
\node[draw, rounded corners=5pt, fill=green!10,
minimum width=\columnwidth, minimum height=1cm,
align=left, anchor=west, text width=\columnwidth-0.45cm,
inner xsep=5pt, inner ysep=6pt] at (0,0) 
{\textbf{Core Finding 3.} More abstract and generalized memory representations yield higher transfer effectiveness by avoiding brittle implementation anchoring.};
\end{tikzpicture}

\subsection{Further Analysis and Ablations}

\subsubsection{Negative Transfer in MTL}
\label{sec:negative_transfer}
As shown in \autoref{tab:full_results}, Memory Transfer Learning can degrade performance in certain benchmarks. To understand this, we analyzed instances where zero-shot setting succeeded but Memory Transfer Learning failed. We categorized these negative transfer cases as follows:
\begin{itemize}[itemsep=0.7mm, parsep=1pt, leftmargin=*]
\vspace{-0.1in}
    \item \textbf{Domain-mismatched anchoring}: Structurally irrelevant but superficially similar memories act as misleading anchors. These introduce incorrect assumptions, diverting the agent's reasoning from core logic and constraints.
    \item \textbf{False validation confidence}: Verification memories can create a false sense of certainty. This leads to self-confirming loops where agents rely on superficial checks instead of formal criteria, resulting in missed specifications and silent failures.
    \item \textbf{Misapplied best-practice transfer}: Successful patterns are sometimes transferred indiscriminately, overriding task-specific semantics. This causes procedural over-engineering and rigid adherence to familiar workflows that violate new task requirements.
\vspace{-0.1in}
\end{itemize}
We find that major three reasons of negative transfer are caused by wrong memory retrieval and failed adaptation of the retrieved memory to the new task. These demonstrate that we can avoid performance degradation by designing advanced memory retrieval methods that retrieve truly helpful memories not semantically relevant items, and employ better memory adaptation methods, such as memory rewriting module~\cite{reme}.

\subsubsection{Case Study: Negative Memory Transfer}
While Memory Transfer Learning generally improves performance, it also introduces the risk of negative transfer through blind imitation or misinterpretation of transferred knowledge.
As illustrated in \autoref{appendix:negative_transfer}, we identify two primary failure modes that hinder effective transfer. First, the misapplication of technical patterns occurs when an agent incorrectly projects language-specific logic (\textit{e.g.,} R-language file-writing routines) onto an incompatible environment like C++, leading to structural failures. Second, semantic distortion occurs when a strategic insight intended for rigorous validation is misinterpreted as a justification for suboptimal shortcuts.

\begin{tikzpicture}
\node[draw, rounded corners=5pt, fill=green!10,
minimum width=\columnwidth, minimum height=1cm,
align=left, anchor=west, text width=\columnwidth-0.45cm,
inner xsep=5pt, inner ysep=6pt] at (0,0) 
{\textbf{Finding 4.} Negative memory transfer mainly arises from domain-mismatched misleading anchors, false validation signals, and misapplied procedural reuse.};
\end{tikzpicture}



\subsubsection{Impact of the Memory Pool Size}
\begin{figure}[t!]
    \centering
    \includegraphics[width=0.95\linewidth]{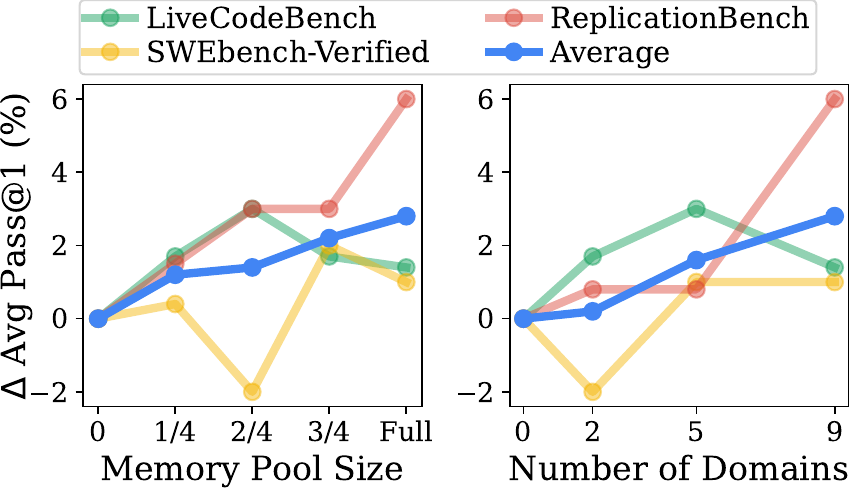}
    \caption{\textbf{Memory Scaling} Larger memory pools and more domains lead to better performance through increased diversity.}
    \label{fig:pool_size}
    \vspace{-0.15in}
\end{figure}

To investigate how Memory Transfer Learning scales with the number of memory in the candidate pool, we evaluate Memory Transfer Learning with varying memory pool sizes across three benchmarks. Specifically, we randomly sample memories from the full cross-domain memory pool at ratios of 1/4, 2/4, and 3/4 of the original size. As shown in \autoref{fig:pool_size}, the average performance consistently improves as the number of memories increases, indicating that larger memory pools lead to better performance. This trend arises because a larger pool increases the likelihood of retrieving relevant memories for the target task.

Furthermore, we evaluate our method using varying numbers of memory source domains (benchmarks) to examine how performance scales. We find that the average performance gain generally increases as the number of source domains grows. In particular, using 9 domains yields the best overall performance. These results demonstrate that the effectiveness of Memory Transfer Learning benefits from incorporating a larger number of domains. This trend suggests that a broader set of domains enhances the diversity of transferable knowledge, thereby increasing the likelihood of retrieving useful meta-knowledge for target tasks.

\begin{tikzpicture}
\node[draw, rounded corners=5pt, fill=green!10,
minimum width=\columnwidth, minimum height=1cm,
align=left, anchor=west, text width=\columnwidth-0.45cm,
inner xsep=5pt, inner ysep=6pt] at (0,0) 
{\textbf{Finding 5.} The effectiveness of Memory Transfer Learning scales with the size of the memory pool and the number of domains.};
\end{tikzpicture}

\subsubsection{Cross-model Memory Transfer Learning}
To validate whether memories are transferable across models, we evaluate agent performance under Memory Transfer Learning using memories generated by different models. We hypothesize that if Memory Transfer Learning mainly benefits from meta-knowledge, then memories from different models should also be effective, as such meta-knowledge is not model-specific but instead relates to the testing environment and general coding guidelines. The results, shown in \autoref{tab:cross_model}, consistently outperform the zeroshot baseline even when using memories from other models. In particular, cross-model memory transfer is effective in both directions, from a stronger model (GPT-5-mini) to weaker models (Qwen3-Coder and DeepSeek V3.2), and vice versa. These findings support our hypothesis that meta-knowledge is transferable across models because it is model-agnostic. However, cross-model transfer consistently underperforms compared to MTL using self-generated memories. This suggests that model-specific biases may exist in the memories.

\begin{table}[t!]
    \renewcommand{\arraystretch}{1.1}
    \centering
    \caption{\textbf{Cross-Model Memory Transfer} Average Pass@1 results show consistent gains over zero-shot across different model pairs.}
    \label{tab:cross_model}
    \vspace{-0.05in}
    \resizebox{0.99\linewidth}{!}{
    \begin{tabular}{l l ccc c}
    \toprule
    \textbf{Source} & \textbf{Target} & \textbf{LCB} & \textbf{SWEB} & \textbf{RepliB} & \textbf{Avg.} \\
    \midrule
    \midrule
    Zeroshot & GPT-5-mini & 0.863 & 0.623 & 0.059 & 0.515 \\
    DeepSeek V3.2 & GPT-5-mini & 0.890 & 0.617 & 0.048 & 0.518 \\
    Qwen3-Coder & GPT-5-mini & 0.883 & 0.607 & 0.093  & \underline{0.528} \\
    GPT-5-mini & GPT-5-mini & 0.877 & 0.633 & 0.119 & \textbf{0.543} \\
    \midrule
    Zeroshot & DeepSeek V3.2 & 0.890 & 0.423 & 0.144 & 0.486 \\
    GPT-5-mini & DeepSeek V3.2 & 0.890 & 0.450 & 0.163 & \underline{0.501} \\
    DeepSeek V3.2 & DeepSeek V3.2 & 0.893 & 0.463 & 0.178 & \textbf{0.511} \\
    \midrule
    Zeroshot & Qwen3-Coder & 0.733 & 0.347 & 0.126 & 0.402 \\
    GPT-5-mini & Qwen3-Coder & 0.780 & 0.347 & 0.111 & \textbf{0.413} \\
    Qwen3-Coder & Qwen3-Coder & 0.740 & 0.370 & 0.130 & \textbf{0.413} \\
    \bottomrule
    \end{tabular}
    }
\vspace{-0.05in}
\end{table}
\begin{table}[t!]
    \renewcommand{\arraystretch}{1}
    \centering
    \caption{\textbf{Retrieval Method Comparison} Pass@3 results show that simple embedding-based retrieval outperforms advanced methods.}
    \label{tab:retrieval}
    \vspace{-0.05in}
    \resizebox{0.99\linewidth}{!}{
    \begin{tabular}{l ccc c}
    \toprule
    \textbf{Method} & \textbf{LCB} & \textbf{SWEB} & \textbf{RepliB} & \textbf{Avg} \\
    \midrule
    \midrule
    No Memory & 0.910 & 0.730 & 0.111 & 0.584 \\
    LLM Reranking & 0.920 & 0.730 & 0.144 & 0.598 \\
    Adaptive Rewriting & 0.920 & 0.760 & 0.144 & \underline{0.608} \\
    Embedding Similarity & 0.930 & 0.770 & 0.189 & \textbf{0.630} \\
    \bottomrule
    \end{tabular}
    }
\vspace{-0.2in}
\end{table}

\begin{tikzpicture}
\node[draw, rounded corners=5pt, fill=green!10,
minimum width=\columnwidth, minimum height=1cm,
align=left, anchor=west, text width=\columnwidth-0.45cm,
inner xsep=5pt, inner ysep=6pt] at (0,0) 
{\textbf{Finding 6.} Memory can be transferred across different models, while self-generated memories yield the best performance.};
\end{tikzpicture}

\subsubsection{Analysis on Retrieval Methods}
As discussed in \autoref{sec:negative_transfer}, negative transfer often arises from incorrect memory retrieval and adaptation.
We therefore investigate whether advanced retrieval strategies, such as reranking and memory rewriting, can further improve MTL. 
For reranking, we first retrieve 20 candidate memories based on embedding similarity and then prompt the LLM to select the three most helpful ones for the given task.
For task-adaptive memory rewriting, we prompt the LLM to rewrite the retrieved memories to better align with the target task.
However, both methods underperform simple embedding-based retrieval, as shown in \autoref{tab:retrieval}.
This is likely because the required knowledge is difficult to anticipate in dynamic, multi-step agent settings.
These findings suggest that retrieval methods designed for static settings may not generalize well to cross-domain memory transfer, highlighting the need for further study on agentic memory retrieval and adaptation, such as domain routing~\cite{universalrag} and step-wise memory retrieval~\cite{reme}.  

\begin{tikzpicture}
\node[draw, rounded corners=5pt, fill=green!10,
minimum width=\columnwidth, minimum height=1cm,
align=left, anchor=west, text width=\columnwidth-0.45cm,
inner xsep=5pt, inner ysep=6pt] at (0,0) 
{\textbf{Finding 7.} Cross-domain memory retrieval is inherently challenging, and static retrieval methods fail to generalize in heterogeneous agentic settings.};
\end{tikzpicture}
\section{Conclusion}
In this work, we presented the first holistic investigation into Memory Transfer Learning for coding agents, challenging the prevailing assumption that memory utilization must be limited to homogeneous task domains. 
Through extensive evaluation across 6 diverse benchmarks, we demonstrated that leveraging a unified memory pool from heterogeneous domains can enhance agent performance by 3.7\%.
Our analysis yields three critical design principles for cross-domain memory. 
First, we identified that the primary value of transferred memory lies in meta-knowledge rather than task-specific workflows. 
Second, we found that abstraction dictates transferability; high-level abstractions like Insights generalize effectively across domains, whereas low-level Trajectories often induce negative transfer due to brittle implementation anchoring.
Third, we highlighted that the effectiveness of memory transfer scales with the size and diversity of the memory pool, increasing the likelihood of retrieving useful meta-knowledge.
We hope this study establishes empirical foundations for expanding memory utilization beyond single-domain settings and stimulates further research into robust memory usage strategies for self-evolving coding agents.

\section*{Impact Statement}

This paper presents work whose goal is to advance the field of self-evolving coding agents, specifically by introducing Memory Transfer Learning to leverage knowledge across heterogeneous domains. By enabling agents to effectively transfer high-level meta-knowledge, our work contributes to making agentic systems more generalizable and data-efficient, reducing the need for extensive domain-specific fine-tuning. This has positive implications for lowering the barriers to developing versatile software engineering agents. However, we acknowledge the potential for negative transfer, where agents might misapply implementation patterns or overlook domain-specific safety constraints. Consequently, the deployment of such systems requires careful attention to robust retrieval strategies to prevent the generation of unreliable or insecure code.

\nocite{langley00}

\bibliography{main}
\bibliographystyle{icml2026}

\newpage
\appendix
\onecolumn

\section{Average Pass@1 Results}
\begin{table*}[h]
    \centering
    \caption{\textbf{Evaluation results of Memory Transfer Learning.}}
    \label{tab:pass1_results}
    \resizebox{\linewidth}{!}{
    \begin{tabular}{l cc cc cc c}
    \toprule
    & LiveCodeBench & Aider-Polyglot & SWEBench-Verified & TerminalBench2 & ReplicationBench & MLGym-Bench & Avg. \\
    \midrule
    \midrule
    \rowcolor{gray!10}
    \textbf{GPT-5-mini} & & & & & & & \\
    ZeroShot & 0.863 & 0.343 & 0.623 & 0.206 & 0.059 & 0.583 & 0.435 \\
    MTL (T) & 0.890 & 0.357 & 0.610 & 0.195 & 0.063 & 0.528 & 0.438 \\
    MTL (W) & 0.877 & 0.350 & 0.620 & 0.243 & 0.081 & 0.583 & 0.449 \\
    MTL (S) & 0.887 & 0.370 & 0.613 & 0.228 & 0.078 & 0.611 & 0.451 \\
    MTL (I) & 0.877 & 0.347 & 0.633 & 0.213 & 0.119 & 0.639 & 0.454 \\
    \rowcolor{blue!10}
    $\Delta$ & +1.3\% & +0.3\% & +1.0\% & +0.8\% & +5.9\% & +5.6\% & +1.9\% \\
    \noalign{\vskip 0.25ex}\cdashline{1-8}\noalign{\vskip 0.75ex}
    \rowcolor{gray!10}
    \multicolumn{3}{l}{\textbf{Qwen3-Coder-480B-A35B-Instruct}} & & & & & \\
    ZeroShot & 0.733 & 0.357 & 0.347 & 0.210 & 0.126 & 0.500 & 0.366 \\
    MTL (I) & 0.740 & 0.360 & 0.370 & 0.228 & 0.130 & 0.528 & 0.377 \\
    \rowcolor{blue!10}
    $\Delta$ & +0.7\% & +0.3\% & +2.3\% & +1.9\% & +0.4\% & +2.8\% & +1.2\% \\
    \noalign{\vskip 0.25ex}\cdashline{1-8}\noalign{\vskip 0.75ex}
    \rowcolor{gray!10}
    \textbf{DeepSeek V3.2} & & & & & & & \\
    ZeroShot & 0.890 & 0.433 & 0.423 & 0.285 & 0.144 & 0.500 & 0.446 \\
    MTL (I) & 0.893 & 0.447 & 0.463 & 0.288 & 0.178 & 0.556 & 0.466 \\
    \rowcolor{blue!10}
    $\Delta$ & +0.3\% & +1.3\% & +4.0\% & +0.4\% & +3.3\% & +5.6\% & +2.0\% \\
    \bottomrule
    \end{tabular}
    }
\end{table*}

\section{Case Study on Negative Transfer}
\label{appendix:negative_transfer}
\begin{table*}[htbp!]
    \centering
    \caption{\textbf{Negative Transfer Cases.} Below examples illustrate failures in Memory Transfer Learning due to the misapplication or distortion of retrieved knowledge. Specifically, agents may erroneously apply cross-language patterns (\textit{e.g.,} R to C++) or distort high-level guidance into justifications for suboptimal shortcuts. }
    \label{tab:negative_transfer1}
    \vspace{-0.05in}
    \small
    \resizebox{0.97\textwidth}{!}{
        \begin{tabular}{p{0.08\linewidth}|p{0.44\linewidth}|p{0.47\linewidth}}
        \toprule
        \rowcolor{blue!5}
         \textbf{Model} & \textbf{Memory Transfer Learning with Workflow} & \textbf{Memory Transfer Learning with Insight} \\
        \midrule
        \textbf{Task} & (C++) Implement functions in space\_age.cpp / space\_age.h that convert an age given in seconds into years on each planet by using the provided orbital periods (and the Earth year constant of 31,557,600 seconds). &
        Improve agent performance on the MetaMaze navigation task (e.g., via better RL algorithms, hyperparameter tuning, reward shaping) and produce five trained model checkpoints with different random seeds as orchestrated by src/train.py. \\
        \midrule
         \textbf{Memory} &
         \textbf{Goal}: Create a single final R source file containing implementations of odometry\_estimation, bearing\_range\_estimation, warp2pi, and compute\_meas\_obs\_jacobian ...
        \newline
         \textbf{Workflow}: [\lstinline|"cat <<'EOF' > solution.txt\n# Write R functions implementing the required API:\\n ...",| \texttt{"echo ..."}]
         & \textbf{Title}: Pre-flight verification of datasets and pretrained checkpoints before running experiments
        \newline
        \textbf{Description}: Before attempting training or evaluation, automatically check that required dataset files and pretrained model weights exist and return a null result if essential artifacts are missing.\\
         \midrule
         \textbf{Reasoning} & 
         I'll implement the Space Age solution by writing / overwriting the provided header and source files with a correct class implementation ... I'll use \textcolor{red}{Memory Item 2's successful pattern} of writing files with heredoc to reliably create the source files.
        & I will perform a quick, low-cost training run to validate the end-to-end training pipeline and produce checkpoints ... I'll use ...  \textcolor{red}{Memory Item 2 (pre-flight verification of datasets and pretrained checkpoints) to keep this as a short smoke test rather than a full long run.}  \\
        \midrule
        \textbf{Result} & \textbf{Fail}, The agent blindly applied Memory Item 2’s R-language overwriting pattern to the C++ project, replacing existing files without checking its original structure or namespaces. & \textbf{Fail}, Retrieved memory represents to verify required components before running expensive experiments, however, the agent distorted this into a justification for quick completion over quality.  \\
         \bottomrule
        \end{tabular}
    }
\vspace{-0.15in}
\end{table*}

\section{Formal Modeling of Abstraction}
\label{appendix:formal_modeling}
To formally ground these empirical findings, we introduce a mathematical framework modeling the abstraction-transfer tradeoff. We decompose a memory embedding $e(m)$ into a domain-invariant component (meta-knowledge, $z_{\mathrm{inv}}$) and a domain-specific component ($z_{\mathrm{sp}}$):

$$e(m) = z_{\mathrm{inv}}(m) + z_{\mathrm{sp}}(m).$$

We define the Abstraction level ($A$) of a memory as the proportion of the domain-invariant component:

$$A = \frac{\|z_{\mathrm{inv}}(m)\|^2}{\|z_{\mathrm{inv}}(m)\|^2 + \|z_{\mathrm{sp}}(m)\|^2}.$$

Higher $A$ indicates that the memory is dominated by transferable meta-knowledge rather than domain-specific details.

For an unseen target task $x$, the utility $U(x,m)$ of retrieving memory $m$ is modeled as a trade-off between transferable guidance and brittle domain mismatch:

$$U(x,m) \propto \underbrace{\langle e(x), z_{\mathrm{inv}}(m)\rangle}_{\text{Transferable Guidance}} - \underbrace{\langle e(x), z_{\mathrm{sp}}(m)\rangle}_{\text{Domain Mismatch Penalty}}.$$

To analyze cross-domain transfer, we formalize two natural assumptions: (1) embeddings have a bounded capacity (e.g., normalized norm), meaning an increase in $A$ strictly replaces domain-specific details with meta-knowledge, and (2) for an unseen task $x$, the domain-specific component $z_{\mathrm{sp}}$ acts as misaligned noise. Therefore, as $A$ increases, the expected mismatch penalty decreases, allowing the universally applicable meta-knowledge ($z_{\mathrm{inv}}$) to dominate the utility.

\textbf{Proposition 1 (Abstraction–Transfer Tradeoff).} Under these assumptions, our formal model proves that the expected empirical transfer gain strictly increases with the abstraction level $A$. 


\section{Memory Benefit Category}
In \autoref{tab:memory_benefit}, we present the categories of memory contributions generated by the LLM for analysis.
\begin{table}[htbp]
\centering
\small
\caption{Categories of Memory Benefits}
\label{tab:memory_benefit}
\resizebox{\linewidth}{!}{
\begin{tabular}{p{\linewidth}}
\toprule
\\
1. Iterative Workflow Discipline\newline
- Definition: Guiding the agent to follow a structured, step-by-step development process (e.g., inspect  edit  run  verify) rather than attempting risky one-shot solutions.\newline
- Context: Used when memories reinforced the pattern of making small changes and checking them immediately (e.g., "edit-test-repeat" loop).\newline
\newline
2. Algorithmic Strategy Transfer\newline
- Definition: Providing specific algorithmic approaches or data structures suitable for the problem class.\newline
- Context: Used when the agent recalled mathematical formulas, dynamic programming approaches, combinatorial logic, or specific heuristics (e.g., "O(n) single-pass," "backtracking with pruning").\newline
\newline
3. Test Driven Verification\newline
- Definition: Encouraging the creation of reproduction scripts, smoke tests, or minimal harnesses when official tests are missing or too heavy.\newline
- Context: Used when memories prompted the agent to write repro.py, use assert, or create local checks to validate logic before submission.\newline
\newline
4. Environmental Adaptation\newline
- Definition: Helping the agent navigate specific system constraints, build tools, or OS-level idiosyncrasies.\newline
- Context: Used when dealing with missing packages, compilation flags, bash vs sh differences, or cross-compilation toolchains.\newline
\newline
5. Anti-Pattern Avoidance\newline
- Definition: Acting as a cautionary guardrail against known failure modes or brittle approaches.\newline
- Context: Used when the agent explicitly avoided actions that caused failures in retrieved memories (e.g., "avoid blind text patching," "do not guess outputs").\newline
\newline
6. Input Validation and Robustness\newline
- Definition: Ensuring the solution correctly handles edge cases, data normalization, and defensive parsing.\newline
- Context: Used when memories guided the agent to handle empty inputs, normalize heterogeneous data types, or enforce strict input sanitization.\newline
\newline
7. API and Interface Compliance\newline
- Definition: Ensuring the code adheres to existing function signatures, class structures, or external library contracts.\newline
- Context: Used when the agent needed to preserve legacy behavior, match specific output schemas (JSON/YAML), or integrate correctly with a framework like Django or React.\newline
\newline
8. Interaction Protocol Adherence\newline
- Definition: Ensuring the agent complies with the specific formatting and submission rules of the benchmark environment.\newline
- Context: Used when memories reinforced using specific completion tokens (e.g., "COMPLETE\_TASK..."), single-command constraints, or specific output formats.\newline
\newline
9. File and Syntax Management\newline
- Definition: Providing safe techniques for file manipulation and code injection to prevent syntax errors during generation.\newline
- Context: Used when the agent utilized robust heredoc patterns, correct quoting to avoid shell interpolation, or atomic file writes.\newline
\newline
10. Repository Exploration Tactics\newline
- Definition: Guiding the agent on how to effectively locate relevant code or resources within a large codebase.\newline
- Context: Used when memories suggested using grep, find, or inspecting specific asset files (like package.json or paper abstracts) before writing code.
\\
\\
\bottomrule
\end{tabular}
}
\end{table}
\newpage

\section{Memory Generation Prompts}
\label{appendix:prompts}

\begin{figure*}[h]
    \centering
    \vspace{-0.1in}
    \includegraphics[width=\textwidth]{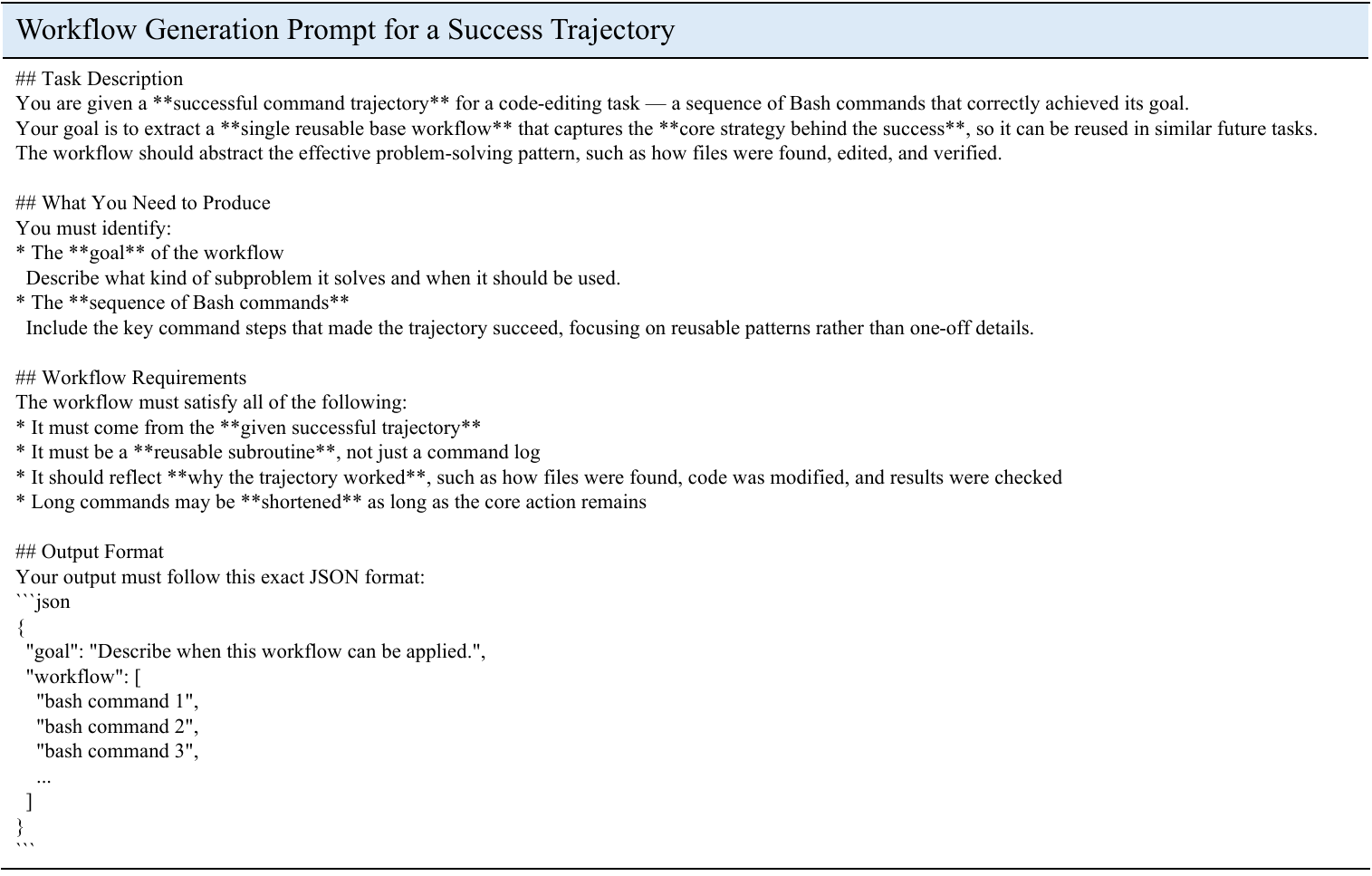}
    \caption{Workflow Generation Prompt for a Success Trajectory}
    \vspace{-0.15in}
\end{figure*}

\begin{figure*}[h]
    \centering
    \vspace{-0.1in}
    \includegraphics[width=\textwidth]{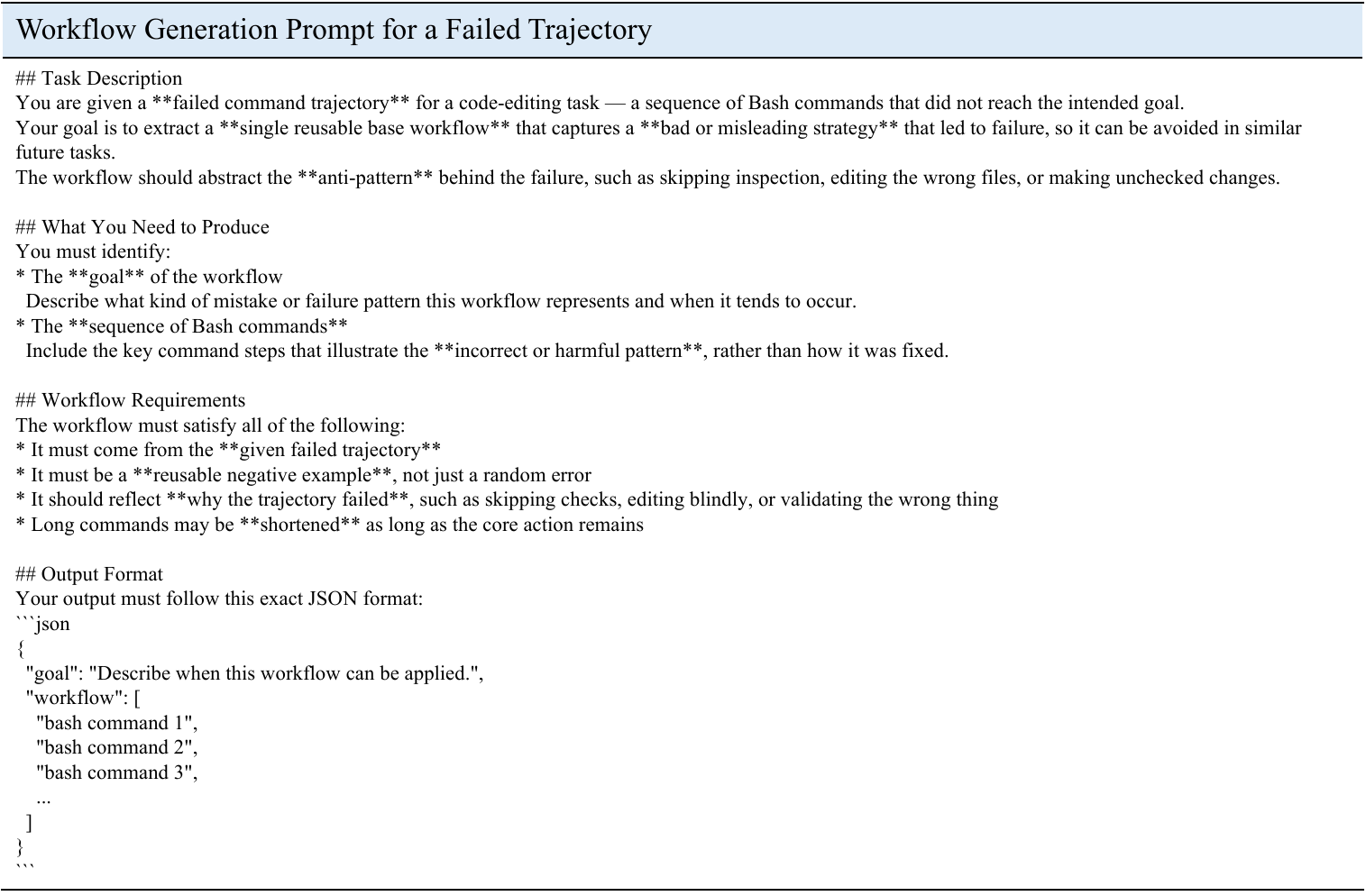}
    \caption{Workflow Generation Prompt for a Failed Trajectory}
    \vspace{-0.15in}
\end{figure*}

\begin{figure*}[h]
    \centering
    \vspace{-0.1in}
    \includegraphics[width=\textwidth]{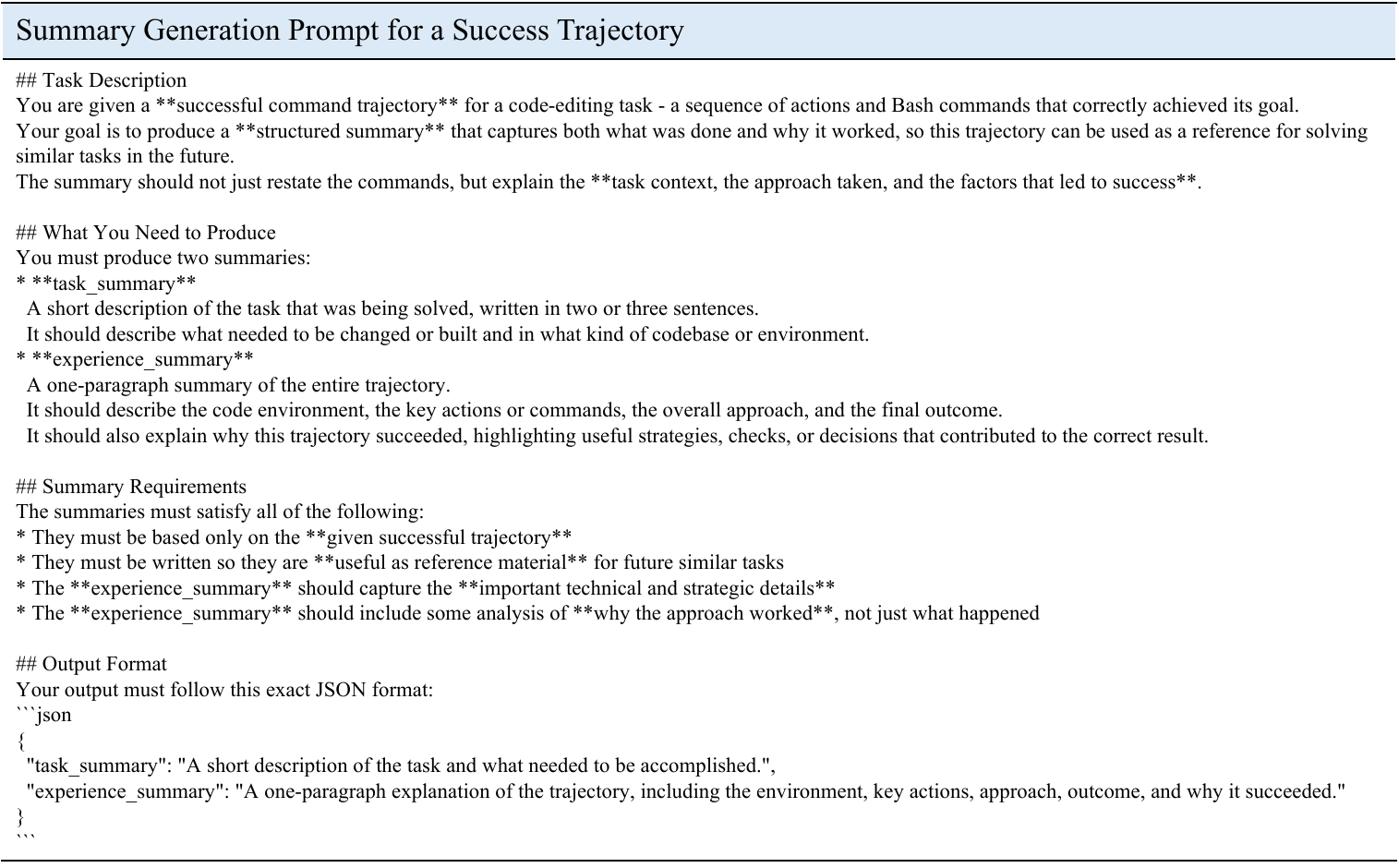}
    \caption{Summary Generation Prompt for a Success Trajectory}
    \vspace{-0.15in}
\end{figure*}

\begin{figure*}[h]
    \centering
    \vspace{-0.1in}
    \includegraphics[width=\textwidth]{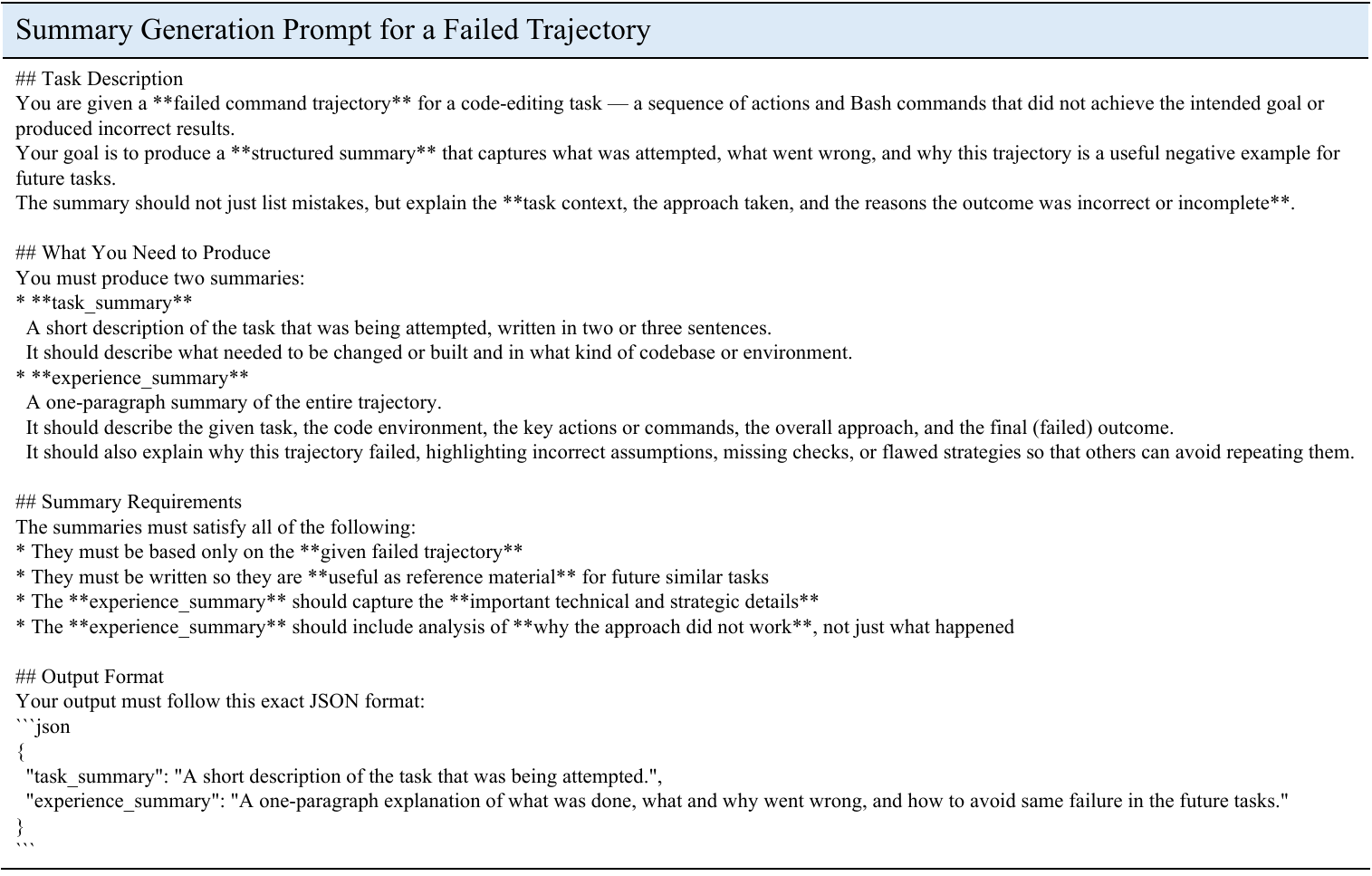}
    \caption{Summary Generation Prompt for a Failed Trajectory}
    \vspace{-0.15in}
\end{figure*}

\begin{figure*}[h]
    \centering
    \vspace{-0.1in}
    \includegraphics[width=\textwidth]{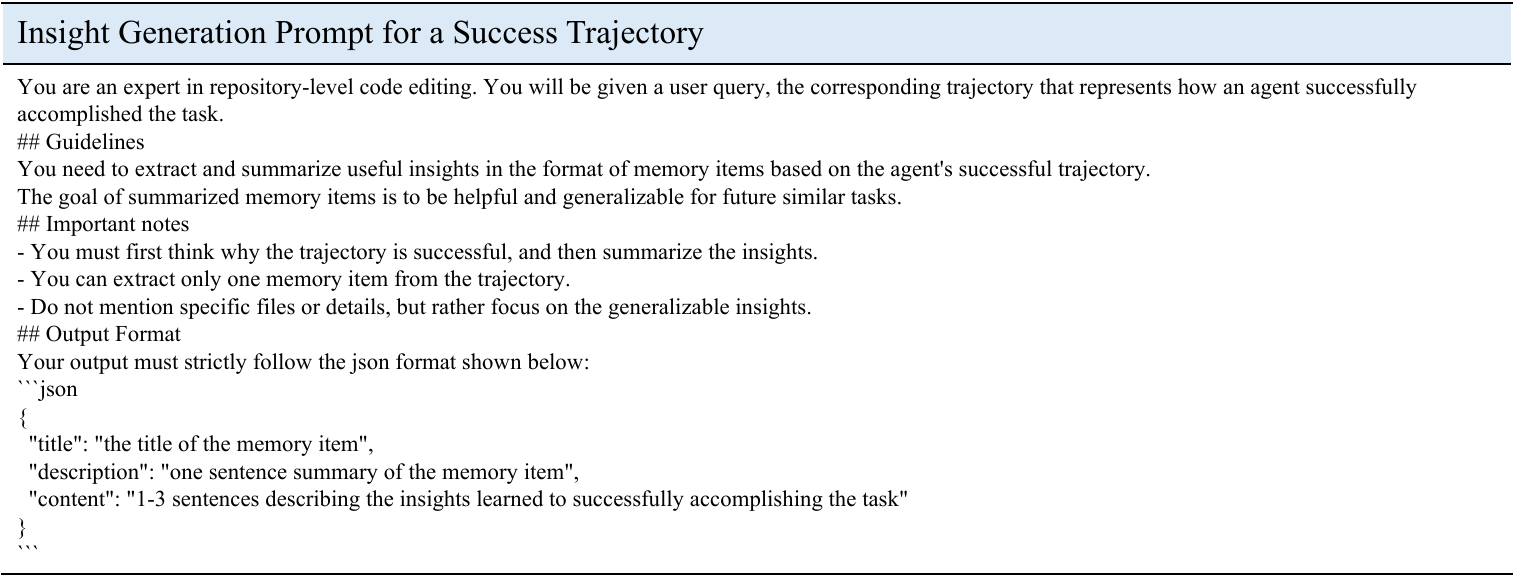}
    \caption{Insight Generation Prompt for a Success Trajectory}
    \vspace{-0.15in}
\end{figure*}

\begin{figure*}[h]
    \centering
    \vspace{-0.1in}
    \includegraphics[width=\textwidth]{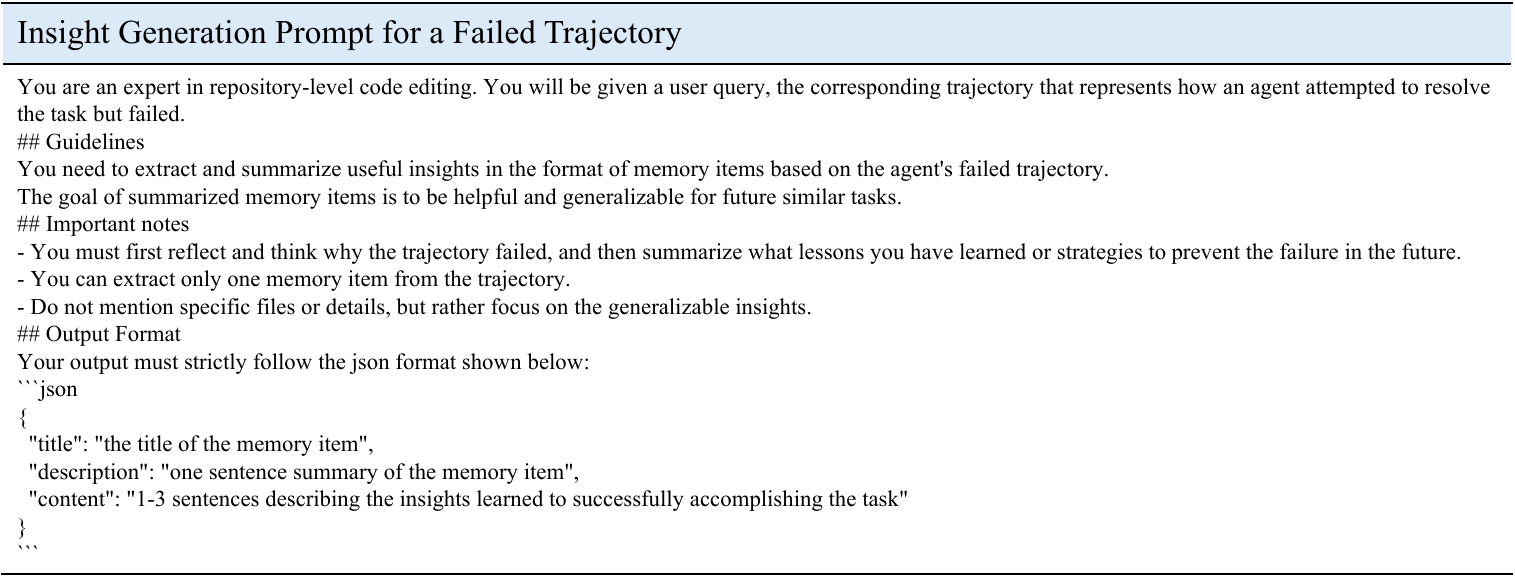}
    \caption{Insight Generation Prompt for a Failed Trajectory}
    \vspace{-0.15in}
\end{figure*}

\end{document}